\documentclass[10pt,twocolumn,letterpaper]{article}

\usepackage[pagenumbers]{cvpr} 

\usepackage[dvipsnames]{xcolor}

\definecolor{cvprblue}{rgb}{0.21,0.49,0.74}
\usepackage[pagebackref,breaklinks,colorlinks,citecolor=cvprblue]{hyperref}
\usepackage{algorithm}
\usepackage{algpseudocode}
\usepackage{xcolor}

\usepackage{makecell} 
\usepackage{multirow}
\usepackage{caption} 

\usepackage{subcaption} 

\def\paperID{207} 
\def\confName{3DV\xspace}
\def\confYear{2026\xspace}

\title{Matrix-free Second-order Optimization of Gaussian Splats with Residual Sampling}

\author{
Hamza Pehlivan \quad 
Andrea Boscolo Camiletto \quad 
Lin Geng Foo \quad 
Marc Habermann \quad \\
Christian Theobalt \quad \medskip\\
Max Planck Institute for Informatics, Saarland Informatics Campus
}
\newcommand{\boldbeta}{\boldsymbol{\beta}}
\newcommand{\JTJ}{\mathbf{J^\top J}}

\begin{document}
\maketitle
%
%
\begin{abstract}
3D Gaussian Splatting (3DGS) is widely used for novel view synthesis due to its high rendering quality and fast inference time.
However, 3DGS predominantly relies on first-order optimizers such as Adam, which leads to long training times.
To address this limitation, we propose a novel second-order optimization strategy based on Levenberg-Marquardt (LM) and Conjugate Gradient (CG), specifically tailored towards Gaussian Splatting.
Our key insight is that the Jacobian in 3DGS exhibits significant sparsity since each Gaussian affects only a limited number of pixels.
We exploit this sparsity by proposing a matrix-free and GPU-parallelized LM optimization.
To further improve its efficiency, we propose sampling strategies for both camera views and loss function and, consequently, the normal equation, significantly reducing the computational complexity. 
In addition, we increase the convergence rate of the second-order approximation by introducing an effective heuristic to determine the learning rate that avoids the expensive computation cost of line search methods. 
As a result, our method achieves a $4\times$ speedup over standard LM and outperforms Adam by $~5\times$ when the Gaussian count is low while providing $\approx 1.3x$ speed in moderate counts.
In addition, our matrix-free implementation achieves $2\times$ speedup over the concurrent second-order optimizer 3DGS-LM, while using $3.5 \times$ less memory.
\\
Project Page: \href{https://vcai.mpi-inf.mpg.de/projects/LM-RS/}{https://vcai.mpi-inf.mpg.de/projects/LM-RS}
\end{abstract}
%
%
%
\section{Introduction} \label{sec:intro}
%
%
%
\begin{figure}[t]
    \centering
    \includegraphics[width=\linewidth]{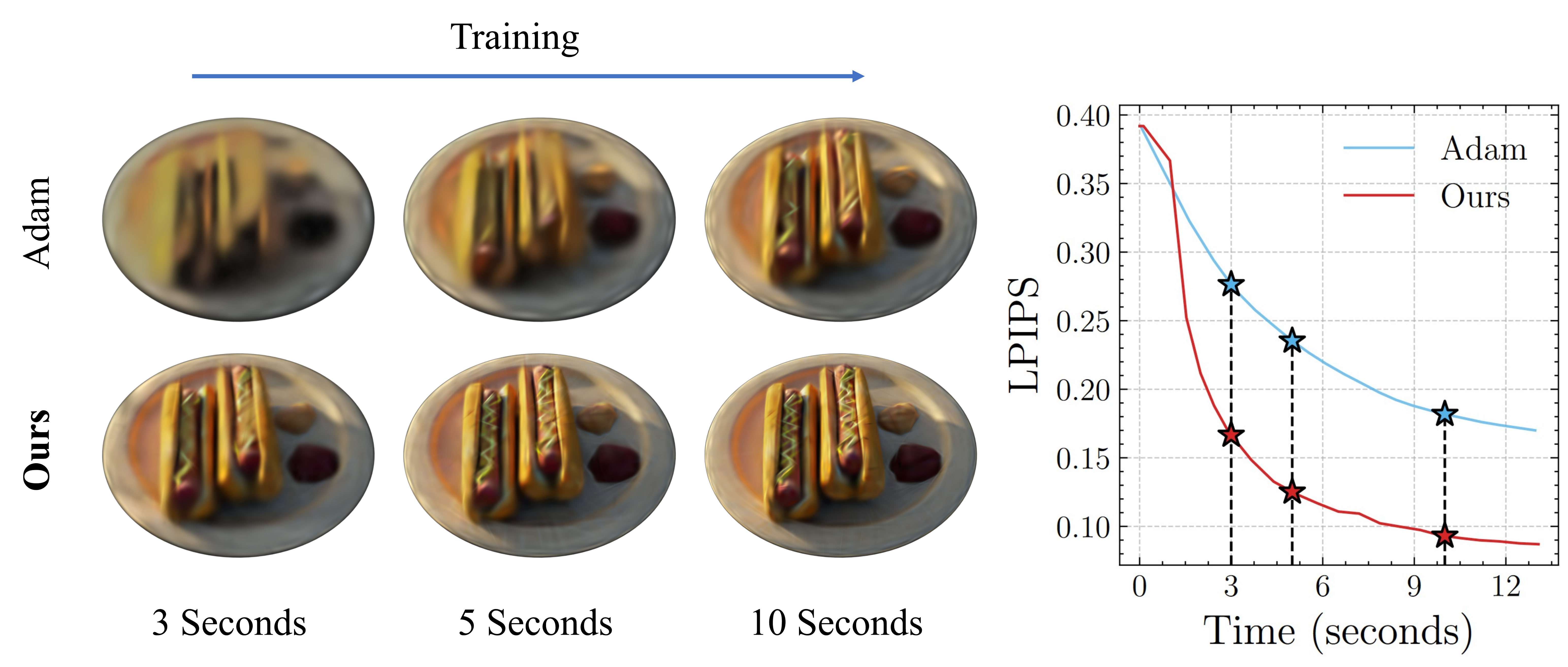}
    \caption{
    We introduce a matrix-free second-order optimizer for Gaussian Splatting. 
    Notably, our dedicated optimizer converges significantly faster than Adam~\cite{adam} and already achieves reasonable renderings after very few seconds of training.
    }
    \label{fig:teaser}
    \vspace{-15pt}
\end{figure}
%
%
%
%
%
Photoreal novel view synthesis from multi-view images or video has attracted significant attention in recent years due to widely applicable downstream tasks in content creation, VR/XR, gaming, and the movie industry, to name a few.
Here, Neural Radiance Fields (NeRF)~\cite{nerf} and 3D Gaussian Splatting (3DGS)~\cite{3dgs} mark a major milestone, due to their unprecedented quality leading to follow ups beyond view synthesis like VR rendering~\cite{VRNeRF}, avatar creation~\cite{ teotia2024gaussianheads}, simultaneous localization and mapping (SLAM)~\cite{matsuki2024gaussianslam, adamkiewicz2022vision}, and scene editing \cite{bao2023sine, chen2024gaussianeditor}. 
%
%
%
\par 
However, NeRF-based models often require substantial training time, and researchers have developed various techniques to mitigate this problem.
Some of them include neural hashing \cite{instantngp}, employing explicit scene modeling \cite{plenoxels}, improved sampling strategies \cite{efficient_nerf}, and tensor factorization methods \cite{chen2022tensorf}.
3DGS~\cite{3dgs} instead does not rely on coordinate-based representations, but leverages a set of 3D Gaussians, which can be effectively rendered into image space using tile-based rasterization.
Nonetheless, optimizing the parameters of each Gaussian can still take hours.
Previous work focused on finding better densification strategies~\cite{taming, mcmc_splatting}, quantization and compression of Gaussians \cite{girish2024eagles, navaneet2023compact3d}, and more efficient implementations for the backward pass \cite{taming}, which led to substantially reduced training times. 
However, all of them mostly rely on a first-order optimization routine, i.e. gradient descent or Adam~\cite{adam}, and do not explore other (potentially second-order) alternatives.
%
%
%
\par 
Second-order optimization is known for having better convergence guarantees compared to first-order methods. 
However, its adoption in 3DGS remains challenging due to the high memory and computational demands associated with storing and inverting large Jacobian matrices.
Notably, the Jacobian size scales quadratically with both the number of Gaussians and image pixels.
Unlike NeRFs, which rely on a dense neural network and a coordinate-based formulation, 3DGS benefits from inherently sparse Jacobians since each Gaussian influences only a small subset of pixels (residuals).
Our \emph{key idea} is to leverage this sparsity in order to make second-order optimization for Gaussian Splatting not only tractable, but also potentially more efficient than first-order baselines. 
%
%
%
\par 
To this end, we formulate fitting the Gaussian parameters to the multi-view images as a non-linear least squares optimization problem and leverage the Levenberg-Marquardt (LM) algorithm for solving it. 
Concurrent work 3DGS-LM \cite{hollein20243dgs} also adopts the LM optimizer with Gaussian Splatting; however, it still suffers from the high storage and computation requirements of the Jacobian matrix. 
In addition, 3DGS-LM has to run the Adam \cite{adam} optimizer first to provide fast convergence, since solving the numerical system arising from the normal equation is computationally expensive.
%
%
%
\par 
To overcome these challenges, we propose a GPU-parallelized matrix-free conjugate gradient solver coupled with pixel sampling, which circumvents explicit storage of the Jacobian matrix and solves the matrix inverse iteratively. 
Firstly, we show that implementing a matrix-free solver naively does not result in a fast optimizer because of the high computational cost of Jacobian-vector products. 
Therefore, we propose to approximate the full normal equation by an effective view sampling strategy and by sampling individual pixels, which results in significantly faster convergence. 
These approximations make the Jacobian-vector products cheaper to begin with and obviate the need for the Adam optimizer.
We also introduce a heuristic to automatically determine the learning rate, which eliminates the need for line search algorithms that are commonly used in conjunction with second-order optimizers. 
%
%
%
\par 
Overall, with these improvements, our proposed second-order optimizer is both memory and compute efficient compared to 3DGS-LM \cite{hollein20243dgs}, suggesting that second-order optimization for 3DGS is a highly promising direction for further study.
Our optimizer demonstrates improvements in settings with a low and moderate number of Gaussians over first-order optimizers as well. 
In summary, we propose a novel second-order optimizer for 3DGS with the following features:
%
\begin{itemize} 
[nolistsep,leftmargin=*]
\item Formulating Gaussian Splatting as a non-linear least squares optimization problem that is solved using a memory and computationally efficient matrix-free Levenberg-Marquardt and conjugate gradient solver specifically tailored towards 3DGS.
\item A view and importance sampling strategy over the pixels (residuals) to effectively approximate the loss, leading to a significant decrease in computational complexity.
\item 
 An effective heuristic to determine the learning rate, which eliminates the need for expensive line search methods while providing stable convergence for 3DGS optimization.
\end{itemize}
%
%
%
\section{Related Work} \label{sec:related_work}
%
\textbf{3D Scene Representation.}
Neural Radiance Fields (NeRFs)~\cite{nerf} represent the 3D geometry of a scene implicitly using a coordinate-based neural network. 
Subsequently, several variations of NeRFs have been proposed to improve visual quality~\cite{barron2021mipnerf} and to improve computational efficiency~\cite{sun2022direct,hedman2021baking,muller2022instant}.
However, despite the significant advancements, NeRFs still face limitations, particularly in terms of their training and rendering efficiency.
More recently, 3D Gaussian Splatting (3DGS)~\cite{3dgs} has been proposed, which employs a fully explicit representation, unlocking significant enhancements in rendering speed while maintaining better qualitative results than NeRF-based methods.
Recent advancements have further improved the rendering quality of 3DGS \cite{lu2024scaffold,yu2024mip}, or its efficiency \cite{fan2025lightgaussian,lee2024compact,fang2024mini,rota2024revising,taming}.
Notably, LightGaussian~\cite{fan2025lightgaussian} and C3DGS~\cite{lee2024compact} focus on pruning Gaussians, directly leading to improvements in rendering speed. 
Another related approach is to constrain the densification process, efficiently modeling the scene with less Gaussians, such as in Mini-Splatting~\cite{fang2024mini} and 3DGS-MCMC~\cite{mcmc_splatting}.
Some other works aim to compress or quantize Gaussians \cite{girish2024eagles, navaneet2023compact3d}.
Taming3DGS~\cite{taming} steers the densification process in a controlled manner, while also devising a parallelized implementation for more efficient backpropagation in CUDA.
Orthogonally to these works, we explore second-order optimizers for 3DGS optimization, which is challenging due to the high memory and computational demands associated with storing and inverting large Jacobian matrices. 
To achieve this, we exploit the sparsity of the Jacobian matrices of 3DGS.
%
%
\par \textbf{First-order Optimizers.}
Stochastic Gradient Descent (SGD)~\cite{sgd} was developed as a general optimization method and has been widely used in modern multi-dimensional deep learning tasks.
With the addition of momentum~\cite{sgd_momentum}, SGD can overcome local optima, a crucial feature in stochastic settings.
However, a key limitation is that SGD applies the same update to all dimensions, which makes it difficult to adapt to complex problems.
To address this, researchers have developed various preconditioners for the gradient information.
Adagrad~\cite{adagrad} accumulates the squares of the gradients and preconditions the gradient with the square root of the accumulated values.
RMSprop~\cite{rmsprop} replaces accumulation with an exponentially moving average, which leads to more stable training.
Adam~\cite{adam}, arguably the most popular optimizer today, combines momentum with the moving average of second-order moments. 
It has also become the default optimizer for many 3D tasks, including NeRF and 3DGS, and provides a strong baseline.
However, despite the efficacy of first-order optimizers, there is strong incentive to explore second-order optimizers due to their attractive properties, as discussed next.
%
%
\par \textbf{Second-order Optimizers.}
Second-order optimizers often approximate the inverse of the Hessian matrix to precondition the gradient, and can offer significant advantages over first-order methods.
They require far fewer iterations because they can obtain the second-order approximation of the loss landscape into account \cite{kfac, martens2010deep, martens2011learning}. 
Additionally, second-order methods are often not as sensitive to learning rate values, and can often estimate it locally through line search algorithms \cite{armijo1966minimization, hollein20243dgs, more1994line}, or trust region methods \cite{conn2000trust, nocedal1999numerical}.
%
%
\par 
A subset of second-order optimizers, particularly those deriving from Gauss-Newton, leverage curvature information to refine parameter updates.
The general update rule for this family of optimizers follows:
\begin{equation}
    \boldbeta_{t+1} = \boldbeta_t - \mathbf{H^{-1}} \mathbf{g}
\end{equation}
where $\mathbf{g} \in \mathbb{R}^{n}$ is the gradient, $\mathbf{H} \in \mathbb{R}^{n \times n}$ is the Hessian of the loss function and $\boldsymbol{\beta} \in \mathbb{R}^{n}$ are the parameters to be updated, and $n$ is the number of parameters.
In a naive implementation, storing the Hessian requires $O(n^2)$ space, and computing its inverse takes $O(n^3)$ time, making it impractical for large-scale optimization tasks.
Instead, a numerical approximation of the inverse can be obtained using iterative solvers like conjugate gradient (CG)~\cite{hestenes1952methods}. 
This method can also eliminate the need to store the Hessian matrix, as CG only requires the results of matrix-vector multiplications \cite{martens2010deep, martens2011learning}. 
In the literature, this class of methods is referred to as Hessian-free or matrix-free optimization.
%
%
\par 
Specifically, within the class of second-order optimizers, the Gauss-Newton (GN) algorithm \cite{lecun2002efficient} is sometimes adopted to approximate the Hessian $\mathbf{H}$ with $\JTJ$, where $\mathbf{J}$ is the Jacobian matrix. 
This method effectively approximates the second-order derivative only using first-order derivatives. 
Notably, the Levenberg-Marquardt (LM) algorithm \cite{more2006levenberg} is an extension over GN, interpolating between GN and gradient descent, resulting in more stable optimization. 
Recently, some works \cite{lan20253dgs,hollein20243dgs} also explored second-order optimizers for 3DGS optimization. 
3DGS\textsuperscript{2}~\cite{lan20253dgs} explores an algorithm based on Newton's method, but makes it computationally tractable by limiting the second-order computations to a small locality.
Concurrent work 3DGS-LM~\cite{hollein20243dgs} employs the LM optimizer with a dedicated cache for intermediate Jacobian entries for fast computation.
However, caching Jacobians results in quadratic memory requirements, as it scales with respect to both pixels and parameters. 
In addition, 3DGS-LM heavily relies on Adam optimizer for performance since processing a large cache is also expensive.
%
%
\par 
In our paper, we also adapt the LM optimizer, proposing a fast GPU-parallelized conjugate solver for 3DGS, which is a matrix-free optimization that avoids explicitly storing the Jacobian matrix and solves the matrix inverse iteratively.
Our LM optimizer exploits the sparsity property within 3DGS representations for computational efficiency by an effective view sampling strategy and by residual sampling, which enables the matrix-free implementation and provides faster convergence.
%
%
%
%
{\setlength{\spaceskip}{0.4em} \section{Background - 3D Gaussian Splatting (3DGS)}} 
3DGS is a point-based representation that can reconstruct 3D scenes with high fidelity.
The 3D scene is represented using a set of Gaussians, where each Gaussian has an optimizable mean $\mathcal{X}$ and covariance $\boldsymbol{\Sigma}$:
\begin{equation}
    G(\mathbf{x}) = e^{-\frac{1}{2} \left(\mathbf{x} - \mathcal{X}\right)^T \boldsymbol{\Sigma}^{-1} \left(\mathbf{x} - \mathcal{X}\right)}
    \label{eq:gaussian-definition}
\end{equation}
To ensure the covariance matrix remains positive-definite, it is decomposed into rotation $\mathbf{R}$ and scale $\mathbf{S}$  matrices:
%
\begin{equation}
    \Sigma = \mathbf{R} \mathbf{S} \mathbf{S^\top} \mathbf{R^\top}.
\end{equation}
%
%
%
%
\begin{figure}[t]
    \centering
    \includegraphics[width=1\linewidth]{Images/model_3dv.pdf}
    \caption{
    Overview of our method is given. 
    We start from randomly initialized Gaussians and gradually refine them with Levenberg-Marquardt optimizer.
    Since dealing with the true Jacobian matrix is costly, we approximate it with a tile-aware sampling algorithm.
    After we solve the normal equations with approximated Jacobians, we update the parameters using a learning rate heuristic. 
    Note that the Jacobians are never materialized in memory, and the normal equation is solved with our matrix-free algorithm.
    }
    \label{fig:pipeline-overview}
    \vspace{-10pt}
\end{figure}
%
%
%
To render a pixel, the Gaussians are first projected onto the 2D image plane, and sorted according to their depth.
A pixel value $\hat{I_i}$ is computed using $\alpha$-blending, which combines the color $c$ and per-pixel opacity $\alpha$ of the projected Gaussians: 
%
\begin{equation}
    \begin{aligned}
        \hat{I_i}= & \sum_{n \leq \mathcal{S}} c_n \cdot \alpha_n \cdot  \prod_{m < n} (1 - \alpha_m) \\
        & \text{, where} \quad \alpha_n = o_n \boldsymbol{\pi}_{\text{cam}} \left(G_n\right)
    \end{aligned}
    \label{eq:rendering-equation}
\end{equation}
%
where $\mathcal{S}$ represents the total number of depth-sorted Gaussians projected onto pixel $i$, and opacity contribution $\alpha_n$ is obtained by multiplying Gaussian base opacity $o_n$ with its 2D projection (via camera projection $\boldsymbol{\pi}_{\text{cam}}$).

The parameters of the Gaussians are updated by optimizing a loss function between the rendered image $\hat{\mathcal{I}}$ and ground-truth image ${\mathcal{I}}$. 
This optimization is done predominantly using the Adam optimizer:
%
\begin{equation}
    \mathcal{L} = \frac{1}{N} \sum_{i=1}^{N} \mathcal{L}_f (\hat{\mathcal{I}}, \mathcal{I}),
    \label{eq:3dgs-loss}
\end{equation}
%
where $\mathcal{L}_f(\cdot)$ is a loss function that quantifies image differences. 
In this paper, we experiment with mean squared error (MSE) and structural similarity (SSIM) \cite{ssim} loss functions. 
Additionally, 3DGS models the view-dependent effects with spherical harmonics. 
In this work, however, we assume each scene has Lambertian surface and disable the corresponding spherical harmonics levels. 
%
%
\vspace{5pt}
\section{Method} \label{sec:method}
We first introduce the LM optimizer in Sec.~\ref{sec:LMOptimizer}, which we adopt in this work. 
Then, we derive why a naive implementation is not feasible for Gaussian Splatting.
To resolve this issue, in Sec.~\ref{sec:preconditioned_cg_solver}, we discuss our matrix-free approach to adapt the LM optimizer. 
In detail, we propose a GPU-parallelized conjugate gradient solver for our second-order 3DGS optimizer, which circumvents explicit storage of the Jacobian matrix, and solves the matrix inverse iteratively.
Next, in Sec.~\ref{sec:view_sampling}, we introduce a new view sampling strategy to effectively approximate the full normal equation, thereby providing reliable update step directions by integrating information from multiple views.
In Sec.~\ref{sec:monte_carlo_estimation}, we present our residual sampling, providing an approximate loss function, which results in significantly faster convergence. 
Lastly, in Sec.~\ref{sec:additional_improvements}, we introduce a heuristic to automatically determine the learning rate, which eliminates the need for line search algorithms that are commonly used in conjunction with second-order optimizers. 
The overview of all the components is given in Fig. \ref{fig:pipeline-overview}.
%
%
\subsection{Levenberg-Marquardt Optimizer for 3DGS} \label{sec:LMOptimizer}
We use the LM optimizer with a fixed damping parameter to compute an update step by minimizing the loss over a mini-batch $B$, which includes images of the shape height $H$, width $W$, and channels $C$. 
The optimizable parameters for each Gaussian are opacity $o \in \mathbb{R}$, color $c \in \mathbb{R}^3$, mean value $ \mathcal{X} \in \mathbb{R}^3$ , scale $ s \in \mathbb{R}^3$ , and quaternion rotation $q \in \mathbb{R}^4$.  
We represent all Gaussian parameters with $\boldbeta \in \mathbb{R}^P$, where we denote the total number of optimizable parameters with $P$.
\par 
The rendering loss function in Eq. \ref{eq:3dgs-loss} can be rewritten as a nonlinear least squares objective: 
%
\begin{equation}
    \mathcal{L}= \frac{1}{M}\sum_{i=1}^{M} (r_i)^2 
    \label{eq:loss-LM}
\end{equation}
where $r$ is a \textit{residual} and $M=BHWC$. 
The residuals are defined as pixel and structural similarity losses; however, to keep the derivation concise, we omit the structural similarity term: 
\begin{equation}
        r_i = \hat{\mathcal{I}_i} - \mathcal{I}_i 
    \label{eq:residuals}
\end{equation}
%
Then, the update vector $\Delta\boldbeta \in \mathbb{R}^P$ is retrieved by solving the following \textit{normal equation}: 
\begin{equation}
    \left(\mathbf{J}^T 
    \mathbf{J}
    + \lambda\mathbf{I} \right)
    \Delta \boldbeta = \mathbf{J}^T \mathbf{r} 
    \label{eq:normaleq-3DGS}
\end{equation}
%
where $\lambda$ is the fixed damping parameter, $\mathbf{r} \in \mathbb{R}^M$ is the vectorized form of the residuals $r_i$, and $\mathbf{J} \in \mathbb{R}^{M \times P} $ is the Jacobian of $\mathbf{r}$. 
%
\begin{algorithm}[t]
\small
\caption{One step of LM optimizer with matrix-free PCG solver.}
\label{alg:LM_PCG}
\textbf{Input}: Gaussians $\boldbeta_k$, cameras $\mathcal{C}$  \\
\textbf{Output}: Updated Gaussians $\boldbeta_{k+1}$
\begin{algorithmic}[1]
\State $\mathcal{I}, \mathcal{C}_{B}= \text{getBatch} (\mathcal{C})$ \hfill \Comment{View Sampling, Sec. \ref{sec:view_sampling}}
\State $\hat{\mathcal{I}} = \text{splatting}({\boldbeta_k}, \mathcal{C}_B)$ 
\State $\mathcal{\mathbf{r}} = \text{getResiduals}(\mathcal{I}, \hat{\mathcal{I}})$ \hfill \Comment{ Eq. \ref{eq:residuals}}
\Statex \hspace*{-\algorithmicindent} \Comment{Beginning of matrix-free PCG solver (Sec \ref{sec:preconditioned_cg_solver}) \, \, \, \, \, \, \, \, \, \, \, \,}
\State $\mathbf{r}_0 = \mathbf{J^\top r}$ \Comment{Estimated  in Sec. \ref{sec:monte_carlo_estimation}}
\State $\mathbf{M}^{-1} = 1 / \text{Diag}(\JTJ + \lambda \mathbf{I})$ \Comment{Estimated  in Sec. \ref{sec:monte_carlo_estimation}}
\State $\mathbf{z}_0 = \mathbf{M}^{-1}\mathbf{r}_0 $
\State $\mathbf{p}_0 = \mathbf{z}_0 $
\State $\mathbf{x}_0 = \mathbf{0} $
\For {$i = 0$ to $\text{PCG}_\text{iters}$}
    \State $\mathbf{u} = (\JTJ + \lambda \mathbf{I}) \mathbf{p}_{i} $  \Comment{Estimated  in Sec. \ref{sec:monte_carlo_estimation}}
    \State $\alpha = \frac{\mathbf{r}_i^T \mathbf{z}_i}{\mathbf{p}_i^T \mathbf{u}}$
    \State $\mathbf{x}_{i+1} = \mathbf{x}_i + \alpha \mathbf{p}_i$
    \State $\mathbf{r}_{i+1} = \mathbf{r}_i - \alpha \mathbf{u}$
    \State $\mathbf{z}_{i+1} = \mathbf{M}^{-1} \mathbf{r}_{i+1}$
    \State $\beta = \frac{\mathbf{r}_{i+1}^T \mathbf{z}_{i+1}}{\mathbf{r}_i^T \mathbf{z}_i}$
    \State $\mathbf{p}_{i+1} = \mathbf{z}_{i+1} + \beta \mathbf{p}_i$
\EndFor
\State $\boldbeta_{k+1} = \boldbeta_{k} + \eta \, x_{i+1}$ \Comment{Dynamic LR Sched., Sec. \ref{sec:additional_improvements}}
\State $\textbf{return } \boldbeta_{k+1}$
\end{algorithmic}
\end{algorithm}

\par 
One iteration of the LM optimizer is completed after we update all the Gaussian parameters $\boldsymbol{\beta}$ with the learning rate $\eta$: 
%
\begin{equation}
    \boldsymbol{\beta}_{(k+1)} = \boldbeta_{(k)} + \eta \cdot \Delta \boldbeta
\end{equation}
%
Note that in the original 3DGS, for each parameter group (opacity, color, mean, scale, and rotation), a different learning rate is used. 
On the other hand, we use a uniform learning rate across parameters because Gauss-Newton-type methods inherently incorporate parameter scaling through the Hessian approximation $\JTJ$. 
In Sec.~\ref{sec:additional_improvements}, we discuss a heuristic to determine the learning rate at every iteration.  
\par 
In practice, naively solving the normal equation is not feasible, especially in large-scale numerical systems.
First, the Jacobians scale quadratically with both the number of Gaussians and the number of pixels, and storing the Jacobians explicitly in memory quickly becomes infeasible. 
For example, using $100$ images at a resolution of $800 \times 800$ pixels as training data, along with $10\,000$ Gaussians to model the 3D scene, would result in Jacobians exceeding $100$ TB in size. 
Even assuming a sparse storage format with a $99\%$ sparsity, the storage requirement would still exceed $1$ TB for a true LM optimizer.
In addition, explicitly inverting the system matrix has O($P^3$) time complexity when implemented naively.
Thus, computing the explicit inverse is often infeasible, and previous work has focused on solving the normal equation iteratively \cite{martens2010deep,martens2011learning, byrd2011use}, but they still remain computationally expensive, as discussed in the next section.
%
%
\subsection{Matrix-free Preconditioned Conjugate Gradient (PCG) Solver for 3DGS} \label{sec:preconditioned_cg_solver}
We first discuss our matrix-free PCG solver to adapt the LM optimizer. 
In this subsection, we present a GPU-parallelized conjugate gradient solver, which does not need to store the Jacobian matrix explicitly, avoiding the memory issues mentioned in Sec. \ref{sec:LMOptimizer}.
To achieve this, we solve Eq.~\ref{eq:normaleq-3DGS} via the preconditioned conjugate gradient algorithm \cite{hestenes1952methods}, which only needs results of Jacobian-vector products, and finds the solution $ \Delta \boldbeta$ iteratively. 
%
%
\par 
To improve the condition number of the linear system in Eq.~\ref{eq:normaleq-3DGS}, we used the Jacobi preconditioner due to its simplicity and effectiveness. 
More specifically, we use $\frac{1}{\operatorname{diag}( \JTJ + \lambda \mathbf{I})} \in \mathbb{R}^P$, as the precondition.
We provide the implementation of the solver in CUDA, together with efficient Jacobian-vector product kernel, which is implemented with forward mode differentiation and dual numbers \cite{baydin2018automatic, wengert1964simple}. 
See the supplementary material for details about the kernel design. 
The pseudocode of the optimizer is given in Alg. \ref{alg:LM_PCG}.
%
%
\par 
Although this optimizer is able to converge to the final solution in a limited number of steps, it is $4\times$ slower than our final method, as shown in Tab.~\ref{table:distribution-comparison}. 
The main reason for this is that the conjugate gradient algorithm needs to run several iterations, therefore, we need to repeatedly compute the $\mathbf{J}^\top \mathbf{J} \mathbf{p}$ product arising on line $10$ of Alg. \ref{alg:LM_PCG}. 
This kind of computational complexity commonly arises when a second-order optimizer is used in large-scale numerical systems, and usually, an approximation method is used for efficiency. 
The common approximation methods are diagonal \cite{becker1989improving, sophia, adahessian, pesky} and block-diagonal \cite{kfac, roux2007topmoumoute, botev2017practical} approximation of the Hessian or $\JTJ$ matrix.
%
\par 
We observe that the Hessian approximation $\JTJ$ in 3DGS does not exhibit a \textit{dominant} diagonal or block-diagonal structure, as illustrated in Fig. \ref{fig:motivation-sampling}. 
This arises because each pixel is rendered through the interaction of multiple Gaussians, as described in Eq.~\ref{eq:rendering-equation}. 
In other words, no single Gaussian independently represents a surface; instead, it relies on the collective contribution of surrounding Gaussians to accurately capture the ground truth.
This leads to many off-diagonal and off-block-diagonal entries in $\JTJ$.
Therefore, the common diagonal and block-diagonal approximations do not provide good approximations.
To this end, we propose to estimate the loss function with our proposed view and residual sampling. The sampling techniques enable the matrix-free implementation as the cost of solving Eq. \ref{eq:normaleq-3DGS} is drastically reduced.
%
%
\subsection{View Sampling} \label{sec:view_sampling}
Second-order methods like LM are typically used in deterministic settings where the full objective is evaluated at every iteration. 
If this is not the case, estimating local curvature can be problematic and become unreliable \cite{byrd2011use}.
This poses a challenge in the 3DGS setting, where the number of views can exceed hundreds, as incorporating all of them at the same time is infeasible. 
Yet, to compute meaningful gradients, we must find an effective way to approximate the full normal equation in Eq.~\ref{eq:normaleq-3DGS}.
%
%
\par 
To this end, we introduce a camera sampling approach that allows us to get a diverse set of views in each batch.
Let $\mathcal{C} = \{c_1, c_2, \dots, c_N\}$ be the set of all cameras. For each camera $c_i$, we form the feature vector:
\begin{equation}
    f_i = [x_i,y_i,z_i,\;{dx}_{i},{dy}_{i},{dz}_{i}]
\end{equation}
where $(x,y,z)$ is the cameras' normalized location and $({dx},{dy},{dz})$ is the viewing direction.
We then run K-Means clustering on $\{f_i\}_{i=1}^N$ to partition the cameras into batch size number of clusters (e.g., 8 clusters created for a batch size of 8). 
A camera is then randomly selected from every cluster and collected in the set $\mathcal{C}_B$, which serves as training input. 
This approach ensures that each batch captures a more balanced and diverse set of views, resulting in a more accurate estimation of the curvature information. 
As evidenced by Tab.~\ref{table:batchsize-comparison}, this method converges to higher scores compared to the random sampling of the cameras.  
%
%
\subsection{Estimation of Loss Function with Residual Sampling} \label{sec:monte_carlo_estimation}
%
%
\begin{figure}[t]
    \centering
    \includegraphics[width=0.99\linewidth]{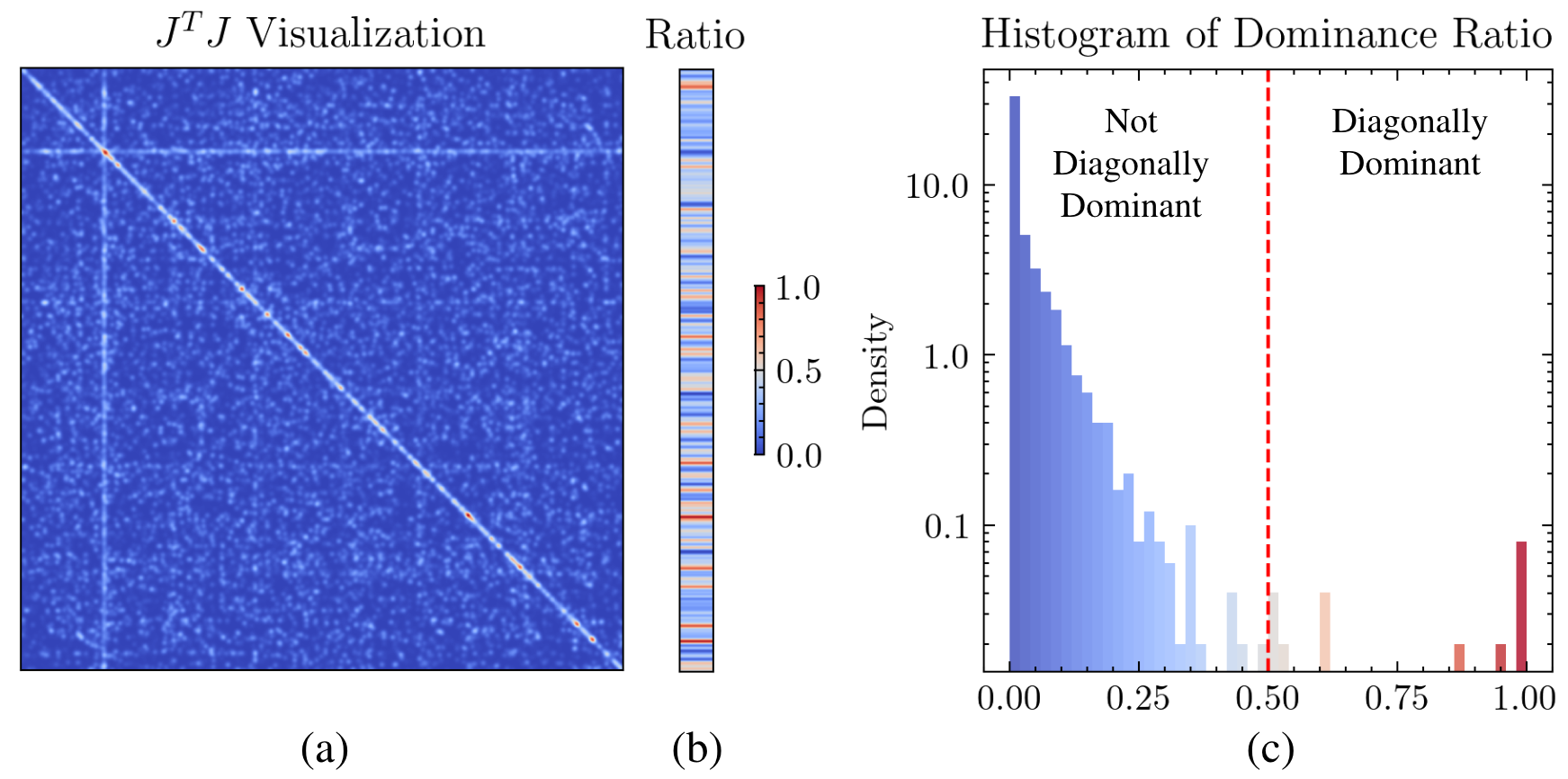}
    \caption{We visualize a normalized $\JTJ$ matrix for one downsampled training image (a). While elements in the diagonal are common, we compute the \textit{dominance ratio} as the normalized ratio between the diagonal element over the sum of other elements in the same row (b) and show how only a limited number of parameters lead to a diagonally dominant linear system (c).
    }
    \label{fig:motivation-sampling}
    \vspace{-15pt}
\end{figure}

%
%
Additionally, to improve the efficiency and feasibility of our second-order optimizer for 3DGS, we further exploit the sparsity property within the 3DGS representation, which has been discussed in Sec.~\ref{sec:preconditioned_cg_solver}.
Yet, at the same time, as observed in Fig.~\ref{fig:motivation-sampling}, there does not exist a well-organized structure to the sparsity.
Instead, in this paper, we propose to adopt a residual sampling approach.  
%
%
\par 
We begin our derivation by rewriting our least squares loss function in Eq.~\ref{eq:loss-LM} as an integral:
Then, the loss function is given as: 
%
\begin{equation}
  L =  \int_{\Omega} r(\mathbf{p})^2 \, d\mathbf{p}
\label{eq:integral-loss}
\end{equation}
%
where the integral is defined over $\Omega$, the union of all image domains of all cameras, and calculated at the coordinates  $\mathbf{p} \in \mathbb{R}^2$ of the image plane. 
To estimate the loss function $L$, we express it as an expectation with respect to an arbitrary probability distribution $q(\mathbf{p})$:
%
\begin{equation}
  L =  \int_{\Omega} r(\mathbf{p})^2 \, \frac{q(\mathbf{p})}{q(\mathbf{p})} \, d\mathbf{p} = \mathbb{E}_{\mathbf{p} \sim q(\mathbf{p})} \left[ \frac{r(\mathbf{p})^2}{q(\mathbf{p})} \right].
\label{eq:loss-expected-value}
\end{equation}
%
This formulation allows us to interpret $q(\mathbf{p})$ as a sampling distribution over the domain $\Omega$. Consequently, we can approximate the expectation numerically using Monte Carlo estimation. In particular, by independently sampling $N$ pixels $\{\mathbf{p}_i\}_{i=1}^N$ from $q(\mathbf{p})$, the loss function can be approximated as:
%
\begin{equation}
  L \approx  \hat{L} = \frac{1}{N} \sum_{i=1}^N r(i)^2 \, \frac{1}{q(i)}.
\label{eq:loss-sampled-form}
\end{equation}
%
The estimated loss function allows us to derive the estimated versions of Jacobian-vector products arising in the PCG solver. 
The derivative of the estimated loss function with respect to a parameter becomes: 
%
\begin{equation}
 \frac{\partial \hat{L} }{\partial \boldbeta_j }
 = \frac{2}{N} \sum_{i=1}^N \frac{\partial r_i}{\partial \beta_j} r_i\,  \frac{1}{q(i)}
\label{eq:gradient}
\end{equation}
%
This gradient estimates the right-hand side of the normal equation given in Eq. \ref{eq:normaleq-3DGS} since $\frac{\partial r_i}{\partial \beta_j} = J_{ij}$.
%
%
\par 
Similarly, we can calculate the Hessian of the estimated loss function given in Eq. \ref{eq:loss-sampled-form}. To do so, we take the derivative of Eq. \ref{eq:gradient} with respect to parameter $\boldbeta_k$: 
%
\begin{equation}
\hat{H}_{jk} = \frac{2}{N} \sum_{i=1}^{N} \left(
  \frac{\partial r_i}{\partial \beta_j}
  \frac{\partial r_i}{\partial \beta_k}
  \;+\;
  r_i \,
  \frac{\partial^2 r_i}{\partial \beta_j \,\partial \beta_k}
\right) \, \frac{1}{q(i)}
\label{eq:hessian-entry}
\end{equation}
%
When we apply the Gauss-Newton approximation to the Hessian matrix, the term with second derivative in Eq.~\ref{eq:hessian-entry} is ignored, and we are left with an estimation of $\JTJ$. 
%
%
\par 
Note that this approximation preserves the symmetry and positive semi-definiteness of $\JTJ$, a property required for the convergence of the conjugate gradient algorithm.
%
%
\par 
In this work, we evaluated a range of candidate distributions $q$ and found that simple uniform distribution works well. 
For a comparison of alternative distributions and their respective results, please refer to the Supplementary.
%
%
%
%
%
%
\par 
In practice, sampling pixels over the image does not efficiently integrate with GPU programming and the backward pass of 3DGS. 
The reason is that an image is divided into $16 \times 16$ tiles and when the pixels are selected randomly, some tiles get more samples than others, causing an unbalanced workload among thread blocks.
Moreover, the random nature of the sampling does not allow fixed thread assignment per tile. 
Therefore, we use a stratified sampling strategy by distributing samples among tiles, and we perform sampling inside them.
In other words, instead of sampling $N$ pixels from the entire image, we sample $N / \mathcal{T}$ pixels from each tile, where $\mathcal{T}$ is the total number of tiles in the image. 
%
%
\par 
The sampling mechanism allows us to achieve $4\times$ speedup over the non-approximated LM optimizer, while maintaining similar performance, as shown in Tab. \ref{table:distribution-comparison}. 
%
%
%
%
\subsection{Dynamic Learning Rate Scheduler} \label{sec:additional_improvements}
When second-order optimizers are applied to large-scale systems, numerical errors can occur, causing the solution vector $\Delta \boldbeta$ to overshoot the true loss landscape.
To determine the optimal learning rate, line search algorithms or trust region methods \cite{armijo1966minimization, hollein20243dgs, more1994line, conn2000trust, nocedal1999numerical, bottou2018optimization} are usually employed.  
However, these methods require additional forward or backward passes, increasing the computational overhead and slowing down the overall algorithm.
Rather than relying on these methods, we estimate the learning rate by constraining the maximum update of color values.
In 3DGS, the color parameters range between $\approx -1.77$ and $\approx 1.77$ due to the level $0$ spherical harmonic coefficient.
This gives us a natural bound for the color parameters. 
We trust the update direction $\Delta \boldbeta$, as long as the resulting color change does not exceed 1. If the change surpasses this threshold, we scale $\Delta \boldbeta$ so that the maximum change remains 1.
Notably, this scaling is applied uniformly across all parameters, regardless of their type (e.g., opacity, color, mean, scale, or rotation).
The effectiveness of this learning rate heuristic is demonstrated in Tab. \ref{table:LR-comparison}, and examples of assigned learning rates are given in the Supplementary. 

%
%
%
%
\begin{figure}[th]
    \vspace{-10pt}
    \centering
    \includegraphics[width=\linewidth]{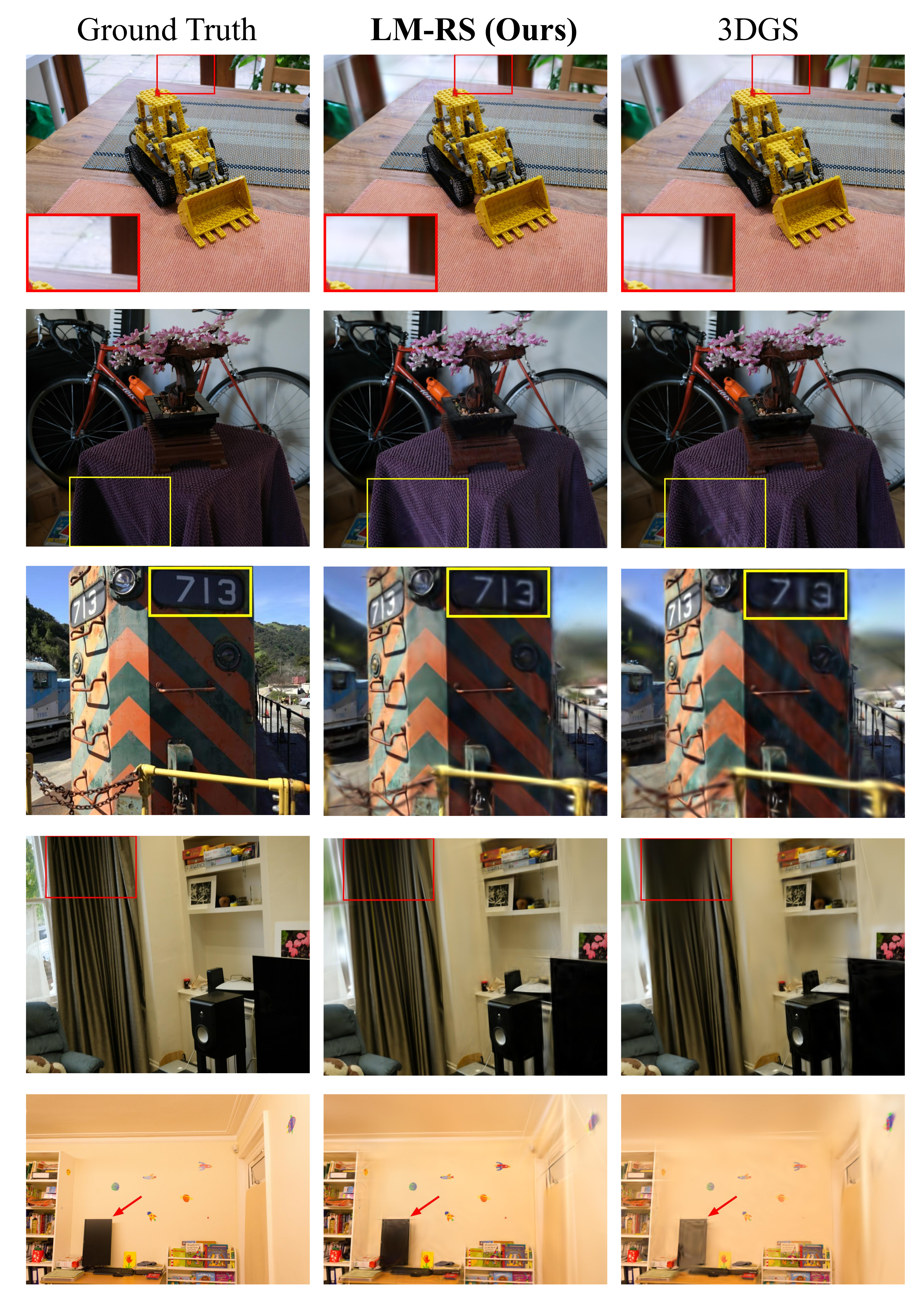}
    \vspace{-20pt}
    \caption{
    We evaluate the performance of the optimizers in the real-world scenes. Our optimizer attains better quality in shorter training time.
    }
    \label{fig:real-world}
    \vspace{-15pt}
\end{figure}
%
%
%
%
\section{Results}  \label{sec:results}
\vspace{-0.7em}
\par \textbf{Datasets and Metrics.} 
We conduct our main experiments on the synthetic NeRF \cite{nerf} and real-world Mip-NeRF360 indoor datasets \cite{barron2021mipnerf}.
%
%
In the synthetic dataset, we position $10{,}000$ Gaussians at random locations with random colors inside of the cube encapsulating the object. 
In the real-world scenes, the Gaussians are initialized with structure-from-motion point clouds as in 3DGS \cite{3dgs}.
For all experiments, we use the default training and test splits. 
We report the test set performance based on structural similarity index (SSIM) \cite{ssim}, learned perceptual image patch similarity score (LPIPS)~\cite{lpips}, and peak signal-to-noise ratio~(PSNR).
%
%
\par \textbf{Baselines.} 
In the synthetic dataset, we compare our optimizer against Adam \cite{adam}, RMSprop \cite{rmsprop}, and SGD with momentum \cite{sgd_momentum}, using their respective implementations in PyTorch.
The gradients are computed using either the vanilla 3DGS or Taming 3DGS \cite{taming}, which enhances the efficiency of the backward computation by storing the gradient state at every $32^{\text{nd}}$ splat. 
\par
In the real-world scenes, we compare against vanilla 3DGS \cite{3dgs} and 3DGS-LM \cite{hollein20243dgs}. For comparisons against 3DGS-LM, we contacted the authors\footnote{At the time of our submission, the code for 3DGS-LM had not yet been released.}, who kindly provided PSNR scores, runtime, and average memory usage across several datasets. 
In addition, they provided results obtained without the Adam optimizer, which are denoted in the tables as 3DGS-LM (w/o Adam).
First-order baselines are run for $10,000$ iterations, and we performed a hyperparameter search to find the best learning rates. 
\par \textbf{Implementation Details.}  \label{par:imp_details}
We conducted synthetic dataset experiments on a NVIDIA Tesla A40 GPU and real-world experiments on a NVIDIA Tesla A100 GPU. 
Our method and the baselines optimize the mean squared error (MSE) loss.
We set the sampled pixels per tile ($N$) to 32 in each experiment.
We use a fixed damping parameter $\lambda$ of $0.1$ for the synthetic NeRF scenes and Mip-NeRF360 dataset.
For the synthetic dataset, the batch size is fixed at $8$; for real-world datasets, it is initialized at $16$ and increased to $32$ after the 50th iteration.
The maximum number of conjugate gradient iterations starts at $3$ for synthetic datasets and at $5$ for real-world datasets, rising to $8$ after the 50th iteration.
%
\begin{table}[ht]
    \centering
    \vspace{-5pt}
    \caption{This table presents the comparisons of competing methods averaged in the Mip-NeRF360 indoor dataset. }
    \vspace{-5pt}
    \resizebox{\linewidth}{!}{
    \begin{tabular}{|l|c|c|c|c|}
    \hline

        \textbf{Method}  & SSIM$\uparrow$ & PSNR$\uparrow$ & Time (s) & Memory (GB) \\
        \hline
        3DGS  & 0.854 & 28.21 & 252 & \textbf{5.81} \\
        3DGS-LM (w/o Adam) & - & 26.72 & 1186 & $\approx$ 50 \\
        3DGS-LM  & - & 28.31 & 398 &  $\approx$ 50  \\
        \textbf{LM-RS (Ours)} & \textbf{0.859} & \textbf{28.59} &  \textbf{190} & 14.75 \\
        \hline
    \end{tabular}
    }
    \label{tab:main-results-mipnerf}
    \vspace{-15pt}

\end{table}
%
\begin{figure}[t]
    \centering
    \begin{subfigure}[b]{0.49\linewidth}
        \centering
        \includegraphics[width=\linewidth]{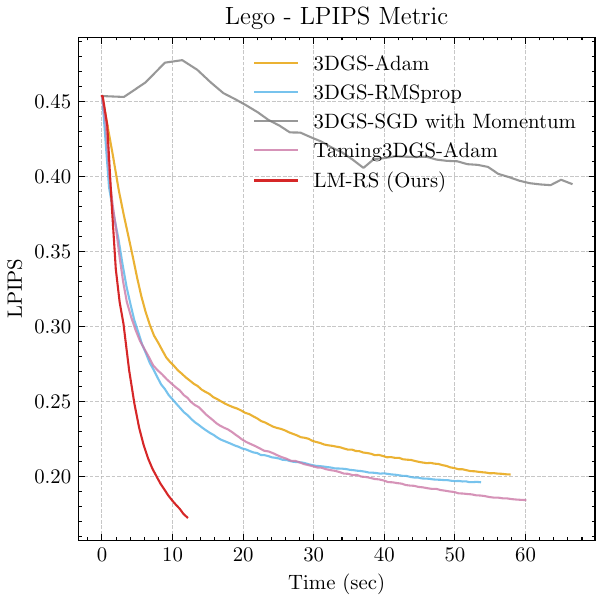}
    \end{subfigure}
    \hfill
    \begin{subfigure}[b]{0.49\linewidth}
        \centering
        \includegraphics[width=\linewidth]{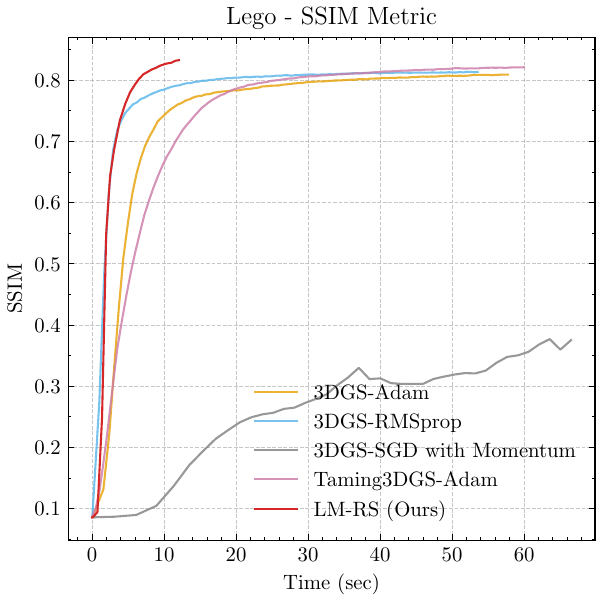}
    \end{subfigure}


    \begin{subfigure}[b]{1.0\linewidth}
        \centering
        \resizebox{\linewidth}{!}{
            \begin{tabular}{|l|c|c|c||c|c|}
            \hline
            \textbf{Method} & LPIPS $\downarrow$  & SSIM $\uparrow$ & PSNR $\uparrow$ & Time (s) & \makecell{ Memory \\ (GB) }  \\
            \hline
            3DGS-SGD &  0.331 & 0.488 &  18.56 &  164 & \textbf{2.43} \\
            3DGS-RMSprop & 0.159 & 0.856 &  25.52 & 54 & \textbf{2.43} \\
            3DGS-Adam &  0.162 & 0.855 &  25.49 & 58 & \textbf{2.43} \\
            Taming3DGS-Adam&  0.147 &  0.862 & 26.01 &  60 & 2.54 \\
            \hline
            \textbf{LM-RS (Ours)} & \textbf{0.139} & \textbf{0.872} & \textbf{26.14} &  \textbf{13} & 3.07 \\
            \hline
            \end{tabular}
        }
        \label{table:mse-results}
    \end{subfigure}
    \caption{The averaged results of competing methods on the synthetic NeRF dataset are presented.  Our method achieves similar quality in a shorter amount of time. We also provide convergence plots on the Lego scene. }
    \label{fig:synthetic-mse}
    \vspace{-15pt}
\end{figure}
\subsection{Comparison}
Next, we evaluate our method, named Levenberg-Marquardt with Residual Sampling (LM-RS), on synthetic NeRF scenes and present the results in Fig.~\ref{fig:synthetic-mse}. 
As shown, our second-order optimizer converges rapidly and achieves a substantial $\approx 5 \times$ speedup over Adam. 
We observed that SGD with momentum consistently underperforms compared to other optimizers, despite our efforts at hyperparameter tuning. 
This highlights the importance of adaptive learning rates for optimizing 3DGS, a feature incorporated in all other evaluated optimizers.
We also did not observe significant speed benefits from the optimizations of Taming 3DGS, which is likely due to a comparatively lower number of Gaussians in this setting that reduces the advantages of gradient caching.
\par
In addition, we provide results in Mip-NeRF360 dataset in Tab.~\ref{tab:main-results-mipnerf} and Fig.~\ref{fig:real-world}.  
Our method  achieves $1.3 \times $ speedup over Adam,  $2 \times $ over 3DGS-LM and $ 6 \times $ speedup over 3DGS-LM (w/o Adam). Thanks to our matrix-free algorithm, we do not rely on the Adam optimizer at all, and our memory requirement is $3.5 \times $ less than 3DGS-LM.    
\par
Please refer to the Supplementary for the results using SSIM loss, as well as quantitative results in Tanks \& Temples \cite{knapitsch2017tanks} and DeepBlending scenes \cite{hedman2018deep}.
%
%
\subsection{Ablation Studies}
\vspace{-0.7em}
In this section, we ablate various design choices incorporated into our optimizer. 
\par \textbf{View Sampling.}
We propose a view sampling approach in Sec.~\ref{sec:view_sampling} to ensure that our optimizer can effectively estimate the local curvature of the loss landscape while still maintaining a relatively low batch size, through sampling of diverse views in a batch. 
In Tab.~\ref{table:batchsize-comparison}, we ablate our design choice, where we observe that our method shows improvements compared to random view sampling.
We also report performance with reduced batch size, where we observe that the performance is significantly lower. 
\begin{table}[h]
\vspace{-5pt}
\scriptsize
\centering
%
%
\caption{The table demonstrates the significance of high batch sizes and the effectiveness of our clustering-based camera sampler compared to random sampler. The results are averaged over Tanks \& Temples scenes. (MBS= Maximum Batch Size)}
\vspace{-4pt}
\resizebox{\linewidth}{!}{
    \begin{tabular}{|l|c|c|c|}
    \hline
    \textbf{View Sampler} & LPIPS $\downarrow$ & SSIM $\uparrow$ & PSNR $\uparrow$ \\
    \hline
    Clustering-Based, MBS=8 &  0.363  & 0.640 & 19.18 \\
    Clustering-Based, MBS=16 &  0.271  & 0.719 & 21.37 \\
    Random, MBS=32 &  0.276 & 0.722 & 21.68 \\
    \hline
    \textbf{Clustering-Based, MBS=32} &  \textbf{0.253}  & \textbf{0.731} & \textbf{21.83} \\
    \hline
    \end{tabular}
}
%
\label{table:batchsize-comparison}
\end{table}
%
\par \textbf{Residual Sampling Size.} 
Our proposed sampling approach in Sec.~\ref{sec:monte_carlo_estimation} is the core algorithm that provides significant speed benefits, enabling faster processing by effectively reducing the dimensionality of highly redundant Jacobians. 
Here, we compare our performance against three baselines in Tab.~\ref{table:distribution-comparison}.
Firstly, when no sampling is used at all, i.e., the full Jacobians are considered every time, the amount of training time required increases significantly, showing the importance of sampling.
When we adopt our pixel sampling, but use a high number of sampled pixels per tile (e.g., $N=128$, $N=64$), there are some gains over no sampling, but its efficiency is still suboptimal.
All these show the efficacy of our residual sampling method.
\begin{table}[ht]
%
%
\centering
\scriptsize
%
%
\caption{We show that sampling fewer pixels per tile (N) has little impact on quantitative performance while substantially increasing efficiency. The results are averaged over the MipNeRF-360 indoor scenes. }
\vspace{-5pt}
\resizebox{\linewidth}{!}{
    \begin{tabular}{|l|c|c|c||c|}
    \hline
    \textbf{Residual Sampler} & LPIPS $\downarrow$ & SSIM $\uparrow$ & PSNR $\uparrow$ & Time (s)\\
    \hline
    LM - No Sampling &  \textbf{0.143}  & \textbf{0.868} &  \textbf{28.68} & 810
    \\
    LM-RS - $N=128$ & 0.159 & 0.860 & 28.63 & 470
    \\
    LM-RS - $N=64$ &  0.160 & 0.860 & 28.66 & 286 \\
    \hline
    LM-RS - $N=32$&  0.161  & 0.859 & 28.59 & \textbf{190} \\
    \hline
    \end{tabular}

}
\vspace{-5pt}

\label{table:distribution-comparison}
\end{table}

%
\par \textbf{Dynamic Learning Rate Scheduler.} 
We compare our learning rate heuristic introduced in Sec.~\ref{sec:additional_improvements} against uniform learning rate, Armijo line search \cite{armijo1966minimization, nocedal1999numerical}, and a grid search method in Tab.~\ref{table:LR-comparison}. 
The grid search method selects the learning rate at each iteration by identifying the value that results in the greatest reduction in the objective function.
Although this method is stable, the learning rates obtained are too pessimistic, and it cannot reach the quality of our heuristic. 
Armijo line search improves the grid search method by picking the highest learning rate that results in \textit{sufficient} reduction of the loss.  
While this approach has been successfully integrated as an improvement in deterministic second-order optimizers, our findings indicate that its effectiveness is limited in stochastic settings, such as ours, where optimization relies on multiple approximations. 
We also found that the uniform learning rate can diverge (LR=0.1) or behave suboptimally (LR=0.07) compared to our scheduler.  
\begin{table}[h]
%
\scriptsize
\centering
%
%
\vspace{-5pt}
\caption{We show that our dynamic learning rate algorithm can converge faster, compared to uniform learning rate or search methods. The results are averaged across Mip-NeRF360 indoor scenes. }
\vspace{-5pt}
\resizebox{\linewidth}{!}{
    \begin{tabular}{|l|c|c|c||c|}
    \hline
    \textbf{Learning Rate} & LPIPS $\downarrow$ & SSIM $\uparrow$ & PSNR $\uparrow$ & Time (s)\\
    \hline
    Uniform $LR=0.07$ &  0.202  & 0.842 & 27.91 & \textbf{188}
    \\
    Uniform $LR=0.1$  & 0.281 & 0.669 & 23.20 & \textbf{188}
    \\
    Grid Line Search  &  0.224 & 0.821& 27.12  & 397
    \\
    Armijo Line Search  &  0.264 & 0.763 & 23.66 & 416 
    \\
    \hline
    Our Scheduler (Sec. \ref{sec:additional_improvements})&  \textbf{0.161}  & \textbf{0.859} & \textbf{28.59} & 190 \\
    \hline
    \end{tabular}
}
\label{table:LR-comparison}
\end{table}

%
%
\vspace{-18pt}
\section{Limitations}
\label{sec:limitations}
We use only a diagonally approximated SSIM loss for performance reasons, which may affect the convergence behavior. Future work could explore more efficient implementations enabling the use of full SSIM loss. In addition, our matrix-free implementation requires more intermediate vectors than Adam, leading $3\times$ higher memory usage.  
\vspace{-5pt}

\section{Conclusion} \label{sec:conclusion}
\vspace{-5pt}
%
%
%
%
\par 
By leveraging the inherent sparsity of the Jacobian matrix and integrating a GPU-parallelized conjugate gradient solver, our method significantly reduces the computational overhead. 
Our novel view and pixel-wise sampling further enhance efficiency, enabling rapid convergence by decreasing the per-step overhead. 
Additionally, our dynamic learning rate scheduler eliminates the need for costly line search procedures, further accelerating training.
Our approach achieves up to $5 \times$ speedup over Adam, particularly excelling in scenarios with low number of Gaussians. 
Overall, our results highlight the potential of second-order methods in accelerating optimization for 3D Gaussian Splatting. 
We anticipate that future work will refine these techniques, particularly by incorporating additional loss terms without incurring runtime overhead, further advancing the efficiency and quality of Gaussian-based scene representations.
%
%

{
    \small
    \bibliographystyle{ieeenat_fullname}
    \bibliography{main}
}
\appendix

\section{Details of CUDA Implementation}
\label{sec:supp-cuda}
\begin{algorithm}[t]
\small
\caption{Forward Pass - 3DGS}
\label{alg:forward-pass}
\textbf{Input}: Sorted Gaussians projected on each pixel: $\mathcal{G}$. \\
\textbf{Output}: Rendered  RGB color C. 
\begin{algorithmic}[1]
\For { \textbf{in parallel} pixel $p$ in \textit{ALL} pixels}
    \State \textbf{float} T = 1.0 \Comment{Initial Transmittance}
    \State \textbf{Vec3} C = 0.0 \Comment{Accumulated Color}
    \For {$g$ in $\mathcal{G}$[$p$]}
        \State // Read Gaussian Properties
        \State \textbf{Vec2} mean2D = $g$.getMean2D()
        \State \textbf{Mat2x2} invCov2D = $g$.getInverseCov2D()
        \State \textbf{float} opac = $g$.getOpacity()
        \State \textbf{Vec3} color = $g$.getColor()
        \State

        \State // Compute the exponential fall-off
        \State \textbf{Vec2} d = mean2D - $p$.getPixelCenter()
        \State \textbf{float} power = $\frac{-1}{2} \cdot d \cdot \text{invCov2D} \cdot d^T$ 
        \State \textbf{float} alpha =  $\text{opac} \times e^{\text{power}} $
        \State
        \State // Update output color and transmittance
        \State C += (alpha $\times$ T )$ \, \cdot \, $ color
        \State T = T $\times \, \text{(1-alpha)}$
    
    \EndFor
 \State $\textbf{return } \text{C}$
\EndFor
\end{algorithmic}
\end{algorithm}

\algblockdefx[CLS]{Class}{EndClass}%
  [1]{\textbf{class} #1}%
  {\textbf{end class}}
  
\algblockdefx[MTH]{Method}{EndMethod}%
  [1]{\textbf{function} #1}%
  {\textbf{end function}}

\begin{algorithm}[t]
\small
\caption{Forward Differentiation - LM-RS}
\label{alg:forward-auto-diff}
\textbf{Input}: Sorted Gaussians projected on each pixel: $\mathcal{G}$. \\
\textbf{Output}: Rendered  RGB color C. 
\begin{algorithmic}[1]
\For { \textbf{in parallel} pixel $p$ in \textit{SAMPLED} pixels}
    \State \textbf{DualFloat} T = (1.0, 0.0) \Comment{Initial Transmittance}
    \State \textbf{DualVec3} C = (0.0, 0.0) \Comment{Accumulated Color}
    \For {$g$ in $\mathcal{G}$[$p$]}
        \State // Read Gaussian Properties
        \State \textbf{DualVec2} mean2D = $g$.getMean2D()
        \State \textbf{DualMat2x2} invCov2D = $g$.getInverseCov2D()
        \State \textbf{DualFloat} opac = $g$.getOpacity()
        \State \textbf{DualVec3} color = $g$.getColor()
        \State

        \State // Compute the exponential fall-off
        \State \textbf{DualVec2} d = mean2D - $p$.getPixelCenter()
        \State \textbf{DualFloat} power = $\frac{-1}{2} \cdot d \cdot \text{invCov2D} \cdot d^T$ 
        \State \textbf{DualFloat} alpha =  $\text{opac} \times e^{\text{power}} $
        \State
        \State // Update output color and transmittance
        \State C += (alpha $\times$ T )$ \, \cdot \, $ color
        \State T = T $\times \, \text{(1-alpha)}$
    
    \EndFor
 \State $\textbf{return } \text{C.prime}$
\EndFor
\State
\Class DualFloat
    \State \textbf{float} real
    \State \textbf{float} prime
    \Method {overload operator$\times$(DualFloat other)}
        \State \textbf{return} DualFloat(real $\times$ other.real, \\ \, \, real $\times$ other.prime + prime $\times$ other.real) \Comment{Product Rule}
    \EndMethod
    \State // Similarly, overload other operators (+, - , /, $e^x$, ...)
\EndClass
\State // Similarly, define DualVec and DualMatrix classes. 
\end{algorithmic}
\end{algorithm}

We follow the parallelization pattern applied in 3DGS \cite{3dgs}, dividing images into $16 \times 16$ blocks, and launching a thread for each pixel.
Our main difference from 3DGS is that our block size is significantly less than $ 16 \times 16 = 256$, due to pixel sampling.
In our implementation, we only support sample sizes that are multiples of $32$, which allows us to do fast warp-level reductions when needed. Now, we describe our CUDA kernels in detail:
\par
\textbf{J$^{\top}$v Kernel.} 
This kernel is similar to the backward pass of 3DGS and is executed in two kernel calls using backward mode differentiation. 
The first kernel computes and forwards the derivatives associated with splatting operations to the Gaussians, while the second kernel computes the derivatives related to Gaussian preprocessing.
The main difference with 3DGS is that the number of threads per block is significantly reduced due to sampling, enabling us to perform faster computations. 
We use warp-level reductions to collect the partial dot products when needed. 
\par
\textbf{Jv Kernel.} This kernel implements forward‐mode differentiation via the dual‐number data structure. 
We overload each operator so that, in a single pass, it performs both the real‐value computation and its derivative.
This approach is more efficient than using backward-mode differentiation when the Jacobian-vector product is needed \cite{baydin2018automatic}. 
\par
Conceptually, this kernel works similarly to the forward pass of 3DGS, but enhanced to carry the derivative information.  
Therefore, unlike J$^{\top}$v kernel, it first carries out Gaussian preprocessing, then implements splatting. 
Since we should additionally carry the derivative information, the shared memory usage is doubled compared to the forward pass of 3DGS. 
%
%
For a comparison between the implementations of the forward pass of 3DGS and our forward-mode differentiation, see Alg. \ref{alg:forward-pass} and Alg. \ref{alg:forward-auto-diff}.
As it can be observed from the algorithm, there is no need for warp reductions in this kernel. 
\par
\textbf{Diag (J$^{\top}$J) Kernel.} To our current knowledge, there is no easy way to split this kernel into two calls. 
Therefore, we create a single kernel that squares and accumulates derivatives using warp-level reductions.
\par
Since the remaining part of the preconditioned conjugate gradient algorithm requires only vector-related computations, it can be efficiently implemented in PyTorch.  
%
%
\section{More Results in Synthetic Dataset}
\label{sec:supp-ssim-loss}
\begin{figure}[ht]
    \centering
    \includegraphics[width=\linewidth]{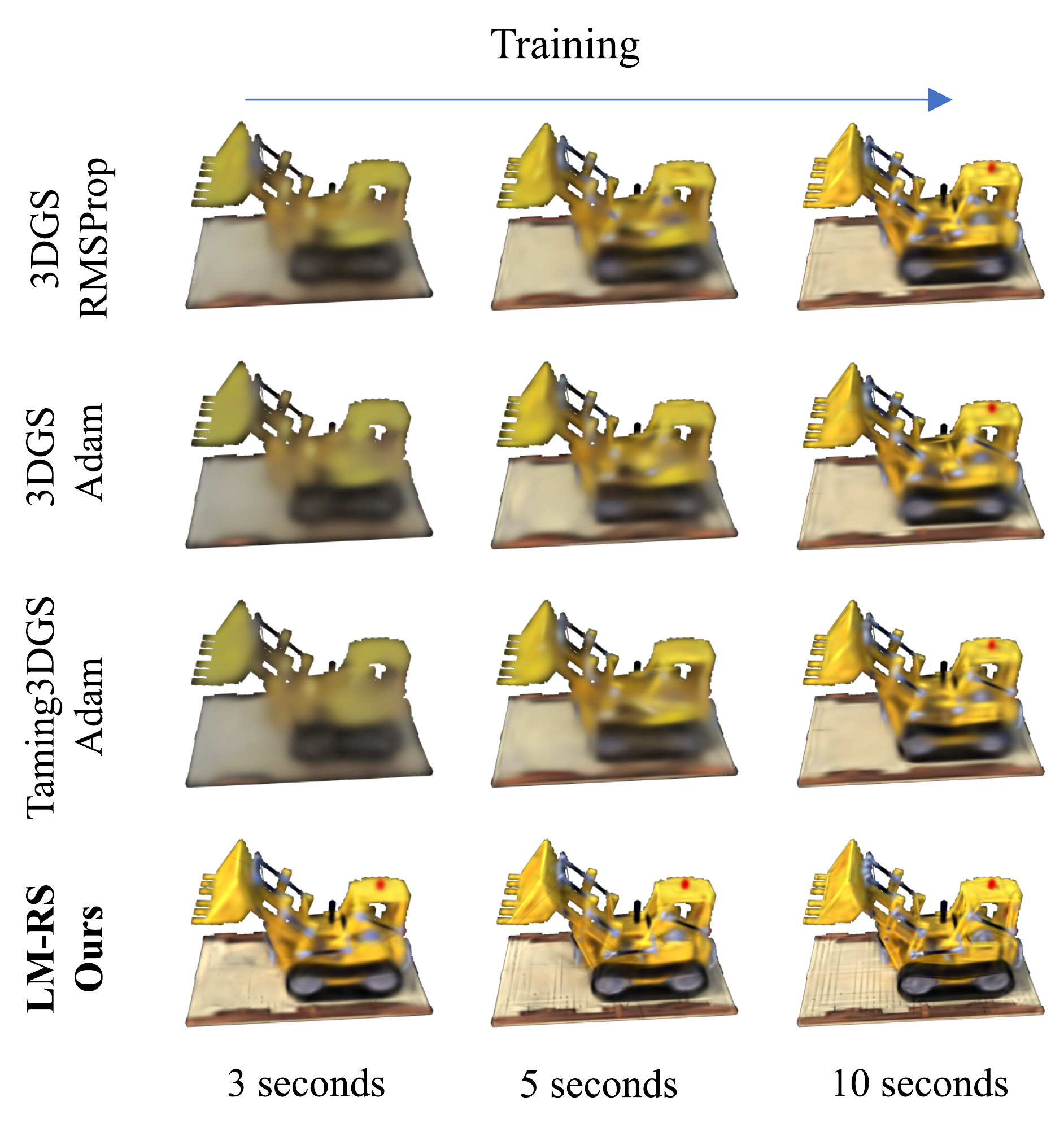}
    \caption{
    We visualize the test set performance of baseline methods evaluated on the Lego scene. Our method achieves comparable metrics in significantly less time. All optimizers use the MSE loss function.
    }
    \label{fig:plots-mse}
\end{figure}

We compare our optimizer against Adam \cite{adam}, RMSprop \cite{rmsprop}, and SGD with momentum \cite{sgd_momentum}, using their respective implementations in PyTorch.
The momentum term of SGD and second-moment factor of RMSprop is set to $0.99$.
For SGD, we used the learning rates $0.16$ for the location, $0.2$ for color and $0.1$ for other parameters.
We also apply a learning rate schedule for the mean value parameters, as suggested in 3DGS.
The gradients are computed using either the vanilla 3DGS or Taming 3DGS \cite{taming}. 
All baselines are run for $10{,}000$ iterations.  
We provide the per-scene scores obtained in the NeRF synthetic dataset~\cite{nerf} in Table \ref{tab:mse-per-scene}. Please also see the supplementary video, comparing Adam and our optimizer in real-time. Additionally, we provide visual results of competing methods in Fig. \ref{fig:plots-mse}.
\par
\par
\textbf{Results with SSIM Loss: } We also conduct experiments using SSIM loss, which is approximated diagonally for performance reasons, as mentioned in the limitations section of the main paper.  
The usual SSIM loss propagates gradients through the local neighbors, which is hard to implement with the parallelization scheme of our $\mathbf{Diag(J^\top J)}$ kernel. 
Our diagonal approximation is similar that used in 3DGS-LM, which follows the derivations in \cite{zhao2016loss}.
In Table \ref{tab:ssim-per-scene}, we show that our optimizer still outperforms the competing methods even with the diagonal approximation.
However, to compensate for this discrepancy, we ran our method for $300$ iterations instead of $200$ iterations. 
The weights are set to $1.0$ and $0.2$, for MSE and SSIM terms respectively. All other hyperparameters remain identical to those specified in the main paper.  
%
%

%

%

%
%
%
%
\section{More Results in Real-world Scenes} \label{sec:supp-realworld}
We report per‐scene results for the Mip‐NeRF360 Indoor, DeepBlending, and Tanks \& Temples datasets in Tab. \ref{tab:results-mipnerf}, Tab. \ref{tab:results-deepblending}, and Tab. \ref{tab:results-tandt}, respectively. 
All optimizers are trained with the mean‐squared error (MSE) loss. 
Notably, our method outperforms both the Adam optimizer and 3DGS‐LM on the Mip‐NeRF360 Indoor and DeepBlending datasets, and beats 3DGS‐LM on Tanks \& Temples. 
\par 
\textbf{Memory Requirements:} As it can be observed from the tables, our method uses more memory than Adam but significantly less memory than 3DGS-LM. 
When updating $n$ parameters, Adam optimizer scales with the number of parameters and uses $O(3n)$ space to store the gradient and two momentum estimates. 
Our matrix-free solver also scales with the number of parameters; however, we need 8 vectors to run the PCG algorithm, making the space complexity $O(8n)$. 
This means both Adam and our optimizer have the same $O(n)$ memory complexity, and our memory requirement differs by only a constant factor. 
On the other hand, 3DGS-LM scales with both the number of pixels ($p$) and the number of parameters, as they need to cache the Jacobian and save the PCG-related vectors, resulting in $O(np)$ space complexity. 
\par \textbf{Results with SSIM Loss: } As mentioned earlier, both 3DGS-LM and our implementation uses a diagonal approximation. In Tab. \ref{tab:results-tandt-ssim}, we compare both methods and show that our optimizer converges faster than 3DGS-LM (w/o Adam). 
However, in this setting, the diagonal approximation becomes overly restrictive, and 3DGS-LM outperforms our method by leveraging the full SSIM loss in the initial steps, which the Adam optimizer conveniently handles.
\section{Details of Dynamic Learning Rate Scheduler} \label{sec:supp-lrsched}
\begin{figure}[h]
    \centering
    \includegraphics[width=\linewidth]{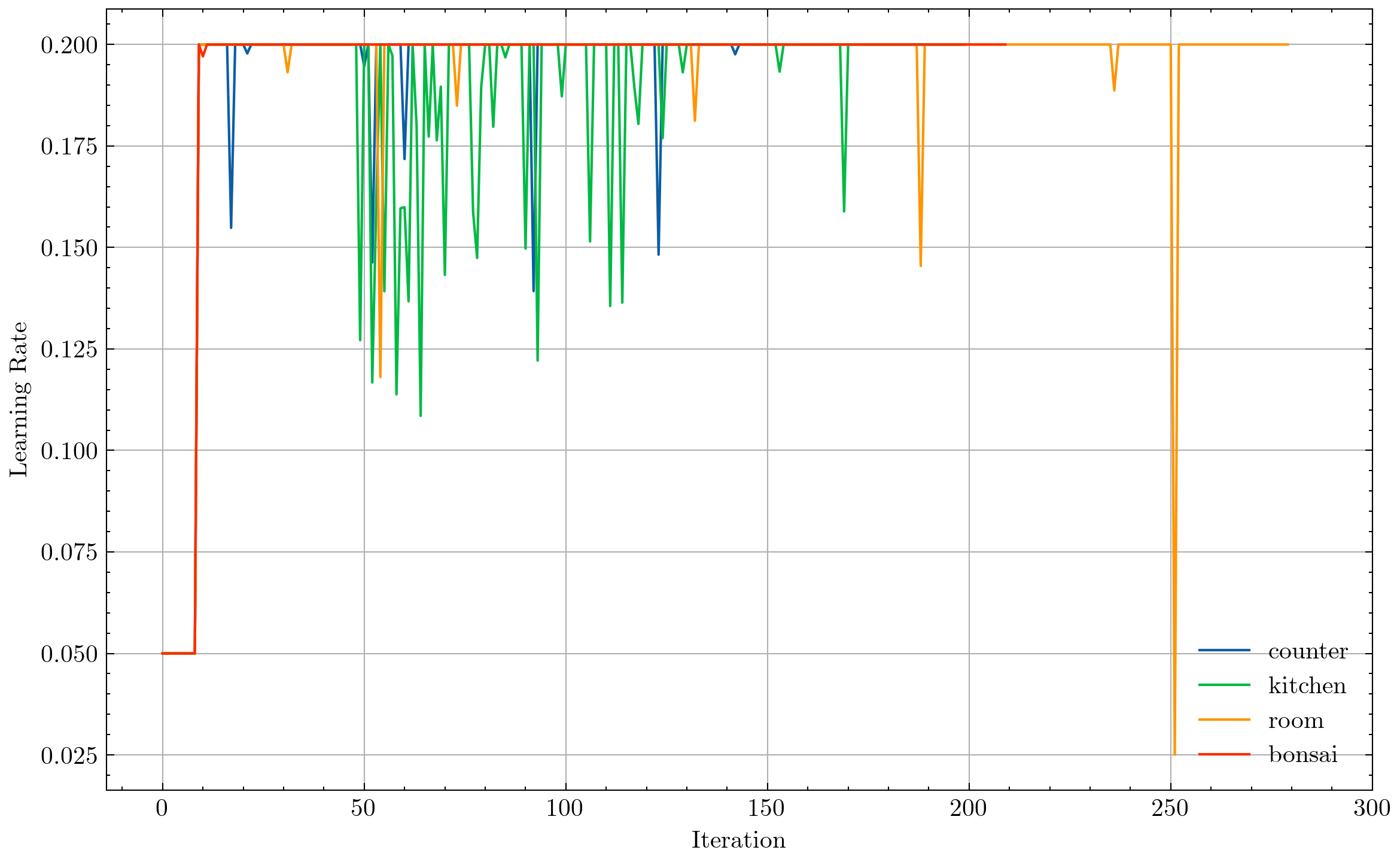}
    \caption{Our dynamic learning rate scheduler can assign both high and low learning rates, resulting in stable and rapid convergence. 
    }
    \label{fig:dynamic-lr}
\end{figure}

Our learning rate scheduler can assign both low and high learning rates, while providing stable training.
We share the obtained learning rates in Mip-NeRF360 scenes in Fig. \ref{fig:dynamic-lr}.
Note that, in all experiments, we cap the maximum learning rate to $0.2$ to avoid overshooting, and we use a fixed learning rate $0.05$ in the first $10$ iterations.
%
%
%
%

%
%

%
\section{Alternative Probability Distributions for Residual Sampling}

We experimented with different pixel (residual) sampling strategies to improve the performance. 
The following candidates are considered for comparison:

$\mathbf{q}_{\textbf{uniform}}$: This distribution assigns equal probability $\frac{1}{M}$ to every pixel $p$, where M is the total number of pixels. This is the distribution used in the main paper.
%
\begin{equation}
   \mathbf{q}_{\text{uniform}}(p) = \frac{1}{M}
\label{eq:uniform-distribution}
\end{equation}
%
$\mathbf{q}_{\textbf{residual}}$: This distribution weights pixel $p$ according to its residual error.
After rendering the Gaussians to obtain the image $\hat{I}$, we compute the per‐pixel residuals against the ground truth image $I$.
We then normalize these residuals using the softmax function: 
\begin{align}
  r_p &= \hat{I}_p - I_p \\
  \mathbf{q}_{\text{residual}}(p) &= Softmax(\left| r_p \right|) = \frac{e^{\left | r_p \right |} }{\sum_k^M e^ {\left| r_k\right|}}
\label{eq:loss-distribution}
\end{align}
$\mathbf{q}_{\textbf{Gaussian}}$: This distribution weights pixel $p$ by the number of Gaussians contributing to it, which we denote as $G_p$.
During the forward pass, we count the number of Gaussians that contribute to pixels' rendering, omitting those that lie beyond the visibility range (i.e., low transmittance) or are transparent. This weight is then normalized by the total number of Gaussians used across all $M$ pixels.
\begin{equation}
    \mathbf{q}_{\text{Gaussian}}(p) = \frac{G_p}{\sum_k^M  G_k}
\end{equation}
\par
We compare the performance of the these distributions in Tab. \ref{table:importance-comparison}.
While other importance sampling methods yield improvements in certain scenes, their average behavior closely matched that of $\mathbf{q}_{\text{uniform}}$. 
Accordingly, we adopt $\mathbf{q}_{\text{uniform}}$ in our paper and conduct all experiments with it.
The similarity in the results could be because of the fixed sample size per tile (32) in every experiment.
Our implementation restricts sample counts to multiples of 32, matching the CUDA warp size to enable efficient warp-level reductions.
Investigating lower sample sizes while avoiding warp divergence can be an interesting direction to explore. 
\begin{table}[h]
%
%
\scriptsize
\centering
%
%
\caption{This table compares the alternative distribution functions for residual sampling. The results are obtained by averaging across Mip-NeRF360 indoor scenes. }
\resizebox{\linewidth}{!}{
    \begin{tabular}{|l|c|c|c|}
    \hline
    \textbf{Distribution} & LPIPS $\downarrow$ & SSIM $\uparrow$ & PSNR $\uparrow$ \\
    \hline
    $\mathbf{q}_{\text{Gaussian}}$&  \textbf{0.160}  & \textbf{0.859} & \textbf{28.65} \\
    $\mathbf{q}_{\text{residual}}$ &  0.161 & 0.857 & 28.49 \\
    \hline
    $\mathbf{q}_{\textbf{uniform}}$ \textbf{(Ours)} &  0.161 & \textbf{0.859} & 28.59 \\
    \hline
    \end{tabular}
}
\label{table:importance-comparison}
\end{table}
\section{Densification of Gaussians}
3DGS proposes a densification algorithm to increase the number of Gaussians, which is disabled in this work. 
The reason is that densification algorithms typically rely on statistics collected over many iterations and complement the dynamics of the Adam optimizer. 
However, second‐order optimizers require far fewer iterations, making it difficult to integrate such algorithms directly. Because of similar reasons, densification is also disabled in 3DGS-LM.
We believe that combining the second-order optimizer with one of the densification strategies would be an interesting research direction.

\begin{table*}[ht]
    \centering
        \caption{Comparisons of different methods on the synthetic NeRF dataset. All optimizers use MSE loss.  }
    \begin{tabular}{l | l | c c c c}
        \toprule
        Method & Scene & LPIPS $\downarrow$ & SSIM $\uparrow$ & PSNR $\uparrow$ & Time (s) \\
        \midrule
         3DGS-RMSprop & \multirow{4}{*}{Chair} & 0.162 & 0.865 & 24.96 & 54 \\
         3DGS-Adam &  & 0.187 & 0.834 & 22.09 & 60 \\
         Taming3DGS-Adam & & 0.145 & 0.876 & 25.75 & 61 \\
         LM-RS (Ours) & & 0.138 & 0.882 & 25.96 & 11 \\
        \midrule
         3DGS-RMSprop & \multirow{4}{*}{Drums} & 0.185  & 0.834 & 22.05  & 52 \\
         3DGS-Adam &  & 0.164 & 0.864 & 24.92 & 55 \\
         Taming3DGS-Adam & & 0.166 & 0.845 & 22.57 & 52 \\
         LM-RS (Ours) & &  0.159 & 0.855 & 22.48  & 10 \\
        \midrule
         3DGS-RMSprop & \multirow{4}{*}{Ficus} & 0.090 & 0.893 & 25.79 & 56 \\
         3DGS-Adam &  & 0.093 & 0.890 & 25.60 & 56 \\
         Taming3DGS-Adam & & 0.078 & 0.903 &  26.51 & 60 \\
         LM-RS (Ours) & & 0.086  & 0.897 & 25.76 & 10\\
        \midrule
         3DGS-RMSprop & \multirow{4}{*}{Hotdog} & 0.111  & 0.923 & 29.65 & 53  \\
         3DGS-Adam &  & 0.111 & 0.924  & 29.78 &  57 \\
         Taming3DGS-Adam & &  0.099 &  0.929 & 30.34 & 60  \\
         LM-RS (Ours) & & 0.085 & 0.936 & 31.04 & 13 \\
        \midrule
         3DGS-RMSprop & \multirow{4}{*}{Lego} & 0.196 & 0.814 & 24.34 & 53 \\
         3DGS-Adam &  & 0.201 & 0.809 & 24.23 & 57 \\
         Taming3DGS-Adam & & 0.184 & 0.821 & 24.82 & 59 \\
         LM-RS (Ours) & & 0.173 & 0.833  & 25.14 & 12 \\
        \midrule
         3DGS-RMSprop & \multirow{4}{*}{Materials} & 0.182 & 0.826 & 23.95 & 51 \\
         3DGS-Adam &  & 0.183  & 0.827 & 23.95 & 56 \\
         Taming3DGS-Adam & & 0.167 & 0.829 & 24.22 & 62 \\
         LM-RS (Ours) & &  0.152 &  0.859 & 24.54 & 10 \\
        \midrule
         3DGS-RMSprop & \multirow{4}{*}{Mic} & 0.104  & 0.922 & 28.14 & 50 \\
         3DGS-Adam &  & 0.108 & 0.920 & 28.09 & 56 \\
         Taming3DGS-Adam & & 0.100 & 0.925 & 28.64 & 63 \\
         LM-RS (Ours) & & 0.090 & 0.930 & 28.56 &  25 \\
        \midrule
         3DGS-RMSprop & \multirow{4}{*}{Ship} &  0.238 & 0.770 & 25.28  & 63 \\
         3DGS-Adam &  & 0.274 & 0.769 & 25.27 & 67 \\
         Taming3DGS-Adam & & 0.238  & 0.767 &  25.44 & 64 \\
         LM-RS (Ours) & & 0.228 & 0.782 & 25.62 & 13 \\
        \midrule
         3DGS-RMSprop & \multirow{4}{*}{ \textbf{Average}} & 0.159 & 0.856 & 25.52 & 54 \\
         3DGS-Adam &  & 0.162 & 0.855 & 25.49 & 58 \\
         Taming3DGS-Adam &  & 0.147  & 0.862 & 26.01 & 60 \\
         \textbf{LM-RS (Ours)} & & \textbf{0.139} & \textbf{0.872} & \textbf{26.14} &  \textbf{13} \\
        \bottomrule
    \end{tabular}

    \label{tab:mse-per-scene}
\end{table*}

\begin{table*}[ht]
    \centering
        \caption{Comparisons of different methods on the NeRF synthetic dataset. All optimizers use a weighted average of MSE and SSIM loss, with coefficients of $1.0$ and $0.2$ respectively. Our method provides $\approx 4\times$ speedup over 3DGS-Adam.  }
    \begin{tabular}{l | l | c c c c}
        \toprule
        Method & Scene & LPIPS $\downarrow$ & SSIM $\uparrow$ & PSNR $\uparrow$ & Time (s) \\
        \midrule
         3DGS-RMSprop & \multirow{4}{*}{Chair} & 0.155 & 0.878 & 24.88 & 85 \\
         3DGS-Adam &  & 0.157 & 0.878  &24.88 & 87 \\
         Taming3DGS-Adam & & 0.138 & 0.890 & 25.66 & 86 \\
         LM-RS (Ours) & & 0.131 & 0.892 & 26.18 & 22 \\
        \midrule
         3DGS-RMSprop & \multirow{4}{*}{Drums} &  0.190 & 0.853 & 21.52  &  83 \\
         3DGS-Adam &  & 0.190 & 0.853 & 21.60 & 86 \\
         Taming3DGS-Adam & & 0.159 & 0.872 & 22.38 & 86 \\
         LM-RS (Ours) & & 0.153 & 0.870 & 22.66 & 21  \\
        \midrule
         3DGS-RMSprop & \multirow{4}{*}{Ficus} &  0.089 & 0.901 & 25.34 & 84 \\
         3DGS-Adam &  & 0.090 & 0.901 & 25.41 & 83 \\
         Taming3DGS-Adam & & 0.074 & 0.913 &  26.28 &  87 \\
         LM-RS (Ours) & & 0.077  & 0.911 & 26.27 & 20 \\
        \midrule
         3DGS-RMSprop & \multirow{4}{*}{Hotdog} & 0.098  & 0.937 & 29.47 & 86 \\
         3DGS-Adam &  & 0.096 &  0.938 & 29.64 & 85  \\
         Taming3DGS-Adam & & 0.087  & 0.942  &  30.29 & 84 \\
         LM-RS (Ours) & & 0.076 & 0.946 & 31.46 & 25 \\
        \midrule
         3DGS-RMSprop & \multirow{4}{*}{Lego} & 0.189 & 0.835 & 24.15 &  83 \\
         3DGS-Adam &  & 0.192 & 0.836 & 24.170 & 86 \\
         Taming3DGS-Adam & & 0.175 & 0.847 & 24.68 & 85 \\
         LM-RS (Ours) & & 0.163 & 0.847  & 25.31 & 23 \\
        \midrule
         3DGS-RMSprop & \multirow{4}{*}{Materials} & 0.175 & 0.864 & 23.84 & 81 \\
         3DGS-Adam &  & 0.176  & 0.863 & 23.80 & 85 \\
         Taming3DGS-Adam&  &  0.159 &  0.870 & 24.12 & 85 \\
         LM-RS (Ours) & &  0.140 &  0.874 & 24.70 & 20 \\
        \midrule
         3DGS-RMSprop & \multirow{4}{*}{Mic} & 0.103 & 0.936 & 27.90 & 81 \\
         3DGS-Adam &  &  0.109 & 0.933 & 27.75 & 80 \\
         Taming3DGS-Adam & & 0.090  & 0.941 & 28.30 & 84 \\
         LM-RS (Ours) & & 0.086 & 0.942 & 28.65 &  26 \\
        \midrule
         3DGS-RMSprop & \multirow{4}{*}{Ship} &  0.227 & 0.804 & 22.77 & 86 \\
         3DGS-Adam &  & 0.225 & 0.806 & 24.90 &  88 \\
         Taming3DGS-Adam &  & 0.214  & 0.811  &  25.15 & 85  \\
         LM-RS (Ours) & & 0.226 & 0.800 & 25.53 & 25 \\
         %

        \midrule
         3DGS-RMSprop & \multirow{4}{*}{\textbf{Average}} & 0.153 & 0.876 & 25.23 & 84 \\
         3DGS-Adam &  & 0.154 & 0.876 & 25.27 & 85 \\
         Taming3DGS-Adam &  & 0.137  & \textbf{0.886} & 25.86 & 85 \\
         \textbf{LM-RS (Ours)} & & \textbf{0.132} & 0.885 & \textbf{26.34} &  \textbf{23} \\
        \bottomrule
    \end{tabular}

    \label{tab:ssim-per-scene}
\end{table*}
\begin{table*}[th]
    \centering
    \caption{Comparisons of competing methods in Mip-NeRF360 Indoor scenes. Our method provides $\approx 1.3 \times$  speedup over Adam,  $\approx 2.1 \times$ over 3DGS-LM, and $\approx 6.2 \times$ over 3DGS-LM (w/o Adam). For 3DGS-LM number of iterations is calculated as: Number of First Order Iterations + Number of Second Order Iterations }
    \begin{tabular}{|l|c|c|c|c|c|c|c|c|c|}
    \hline

        Method & Scene & \# Gaussians& LPIPS$\downarrow$ & SSIM$\uparrow$ & PSNR$\uparrow$ & Time (s) & Memory (GB) & Iteration \\
        \hline
        3DGS & \multirow{4}{*}{counter} & \multirow{4}{*}{ 155767} & 0.190 & 0.822 & 26.63 & 253 & 5.02 & 10,000 \\
        3DGS-LM (w/o Adam) &  &  & - & - & 25.42 & 1332 &  - & 30\\
        3DGS-LM &  &  & - & - & 26.71 & 495 &  - & 10,000 + 10 \\
        \textbf{LM-RS (Ours)} &  &  & 0.189 & 0.829 & 26.83 & 183 & 15.03 & 200 \\
        \hline
        3DGS & \multirow{4}{*}{ kitchen} & \multirow{4}{*}{ 241367}  & 0.136 & 0.857 & 28.03 & 277 & 5.83 & 10,000 \\ 
        3DGS-LM (w/o Adam) & &  & - & - & 26.42 & 1294 &  - & 30  \\
        3DGS-LM & &  & - & - & 28.15 & 536 &  - & 10,000 + 10 \\
        \textbf{LM-RS (Ours)} & &  & 0.134 & 0.864 & 28.61 & 180 &  15.12 & 200 \\
        \hline
        3DGS & \multirow{4}{*}{room}  & \multirow{4}{*}{112627} & 0.176 & 0.858 & 29.52 & 240 & 6.37 & 10,000 \\
        3DGS-LM (w/o Adam) &  &  & - & - & 27.85 & 1087 &  - & 30 \\
        3DGS-LM &  &  & - & - & 29.51 & 271 & - & 10,000 + 5 \\
        \textbf{LM-RS (Ours)} &  &  & 0.177 & 0.861 & 29.61 & 190 & 14.21 & 280 \\
        \hline
        3DGS & \multirow{4}{*}{bonsai} & \multirow{4}{*}{206613} & 0.141 & 0.877 & 28.65 & 240 & 6.04 & 10,000 \\ 
        3DGS-LM (w/o Adam) &  &  & - & - & 27.17 & 1031 &  - & 30 \\
        3DGS-LM &  &  & - & - &  28.85 & 290 &  - & 10,000 + 5\\
        \textbf{LM-RS (Ours)} &  &  & 0.142 & 0.881 & 29.29 & 208 & 14.66 & 210 \\ 
        \hline
        3DGS & \multirow{4}{*}{\textbf{Average}} & ~ & \textbf{0.161} & 0.854 & 28.21 & 252.5 & \textbf{5.81} & \\
        3DGS-LM (w/o Adam) &  &  & - & - & 26.72 & 1186 &  $\approx$ 50 & \\
        3DGS-LM &  &  & - & - & 28.31 & 398 &  $\approx$ 50 & \\
        \textbf{LM-RS (Ours)} &  & ~ & \textbf{0.161} & \textbf{0.859} & \textbf{28.59} & \textbf{190.25} & 14.75 & \\
        \hline
    \end{tabular}
    \label{tab:results-mipnerf}
\end{table*}
\begin{table*}[ht]
    \centering
    \caption{Comparisons of competing methods in DeepBlending scenes. Our method provides $\approx 1.5 \times$  speedup over Adam. }
    \begin{tabular}{|l|c|c|c|c|c|c|c|c|}
    \hline
        Method & Scene & \# Gaussians & LPIPS$\downarrow$ & SSIM$\uparrow$ & PSNR$\uparrow$ & Time (s) & Memory (GB) & Iteration \\ 
        \hline
        3DGS & \multirow{2}{*}{playroom} & \multirow{2}{*}{37005} & 0.205 & 0.850 & 26.90 & 200 & 3.05 & 10,000 \\ 
        \textbf{LM-RS (Ours)} & & & 0.197 & 0.854 & 27.61 & 122 & 6.42 & 400\\
        \hline
        3DGS & \multirow{2}{*}{drjohnson} & \multirow{2}{*}{80861} & 0.275 & 0.840 & 26.74 & 218 & 4.11 & 10,000  \\ 
        \textbf{LM-RS (Ours)} & & & 0.274 & 0.841 & 26.83 & 152 & 8.35 & 400 \\ 
        \hline
        3DGS &  \multirow{2}{*}{\textbf{Average}} &  & 0.240 & 0.846 & 26.82 & 209 & \textbf{3.58} &\\ 
        \textbf{LM-RS (Ours)} &  & & \textbf{0.236} &  \textbf{0.848} & \textbf{27.22} & \textbf{137} &  7.39 &  \\ 
        \hline
    \end{tabular}
    \label{tab:results-deepblending}
\end{table*}
\begin{table*}[ht]
    \centering
    \caption{Comparisons of competing methods in Tanks \& Temples scenes. Our method provides $\approx 1.5 \times$  speedup over 3DGS-LM and $\approx 3.4 \times$ over 3DGS-LM (w/o Adam) optimizer. For 3DGS-LM, the number of iterations is calculated as: Number of First Order Iterations + Number of Second Order Iterations}
    \begin{tabular}{|l|c|c|c|c|c|c|c|c|}
    \hline
        Method & Scene & \# Gaussians & LPIPS$\downarrow$& SSIM$\uparrow$ & PSNR$\uparrow$ & Time (s) & Memory (GB) & Iteration \\ 
        \hline
        3DGS & \multirow{4}{*}{train} & \multirow{4}{*}{182686} & 0.321 & 0.671 & 20.22 & 120 & 2.12 & 10,000\\ 
        3DGS-LM (w/o Adam) &  &  & - & - &  20.09 &	454 & - & 15 \\
        3DGS-LM &  &  & - & - &  20.74 & 172 & - & 10,000 + 5 \\
        \textbf{LM-RS (Ours)} &  &  &  0.298  & 0.687  & 20.22 & 117 & 6.20 & 130 \\
        \hline
        3DGS  & \multirow{4}{*}{truck} & \multirow{4}{*}{136029}  & 0.186 & 0.779 & 23.45 & 117 & 1.78 & 10,000 \\ 
        3DGS-LM (w/o Adam) &  &  & - & - & 23.05 &	291  & - & 10 \\
        3DGS-LM &  &  & - & - & 23.87 &	171  & - & 10,000 + 5\\
        \textbf{LM-RS (Ours)} &  &  & 0.208 & 0.774 & 23.43 & 100 & 5.43 & 130 \\ 
        \hline
        3DGS & \multirow{4}{*}{\textbf{Average}} &  & 0.254 & 0.725 & 21.84 & 118.5 & \textbf{1.95} & \\ 
        3DGS-LM (w/o Adam)&  & & - &  - & 21.57 & 372.5 & $\approx$  40 & \\
        3DGS-LM  &  & & - &  - & \textbf{22.31} & 171 & $\approx$ 40 & \\
        \textbf{LM-RS (Ours)} &  & & \textbf{0.253} &  \textbf{0.731} & 21.83 & \textbf{108.5} &  5.82 & \\
        \hline
    \end{tabular}
    \label{tab:results-tandt}
\end{table*}
\begin{table*}[ht]
    \centering
    \caption{Comparisons of competing methods in Tanks \& Temples scenes. Our method provides $\approx 6.5 \times$  speedup over 3DGS-LM (w/o Adam) while using $6.8 \times$ less memory. The results are obtained using a weight of 1.0 for the MSE loss and 0.25 for the SSIM loss.  For 3DGS-LM, the number of iterations is calculated as: Number of First Order Iterations + Number of Second Order Iterations.}
    \begin{tabular}{|l|c|c|c|c|c|c|}
    \hline
        Method & Scene & \# Gaussians & PSNR$\uparrow$ & Time (s) & Memory (GB) & Iteration \\ 
        \hline
        %
        %
        %
        3DGS-LM (w/o Adam) &  \multirow{3}{*}{train}  &  \multirow{3}{*}{182686} & 19.62 & 604 & - & 30 \\
        3DGS-LM &  &  & 20.52  & 174 & - & 10,000 + 5 \\
        \textbf{LM-RS (Ours)} &  & & 19.99 & 118 & 6.20 & 130 \\
        \hline
        %
        %
        3DGS-LM (w/o Adam) & \multirow{3}{*}{truck} &  \multirow{3}{*}{136029} & 23.05 &	828  & - & 20 \\
        3DGS-LM &  &  & 23.61 &	 173 & - & 10,000 + 5\\
        \textbf{LM-RS (Ours)} &  & & 23.13 & 103 &  5.43 & 130 \\ 
        \hline
        %
        %
        3DGS-LM (w/o Adam)& & & 21.34 & 716 & $\approx$ 40 &  \\
        3DGS-LM  & & & \textbf{22.07} & 173 & $\approx$ 40 &  \\
        \textbf{LM-RS (Ours)} &  & & 21.56 & \textbf{110} & \textbf{5.82} & \\
        \hline
    \end{tabular}
    \label{tab:results-tandt-ssim}
\end{table*}

\end{document}